\documentclass{article}
\usepackage{spconf,amsmath,graphicx}
\usepackage{multirow,soul,xcolor,boldline,hhline,hyperref, enumitem,wrapfig, cite}
\usepackage{pifont}
\usepackage{graphicx,multirow,soul,boldline,hyperref,subcaption,xcolor,soul,flushend}
\usepackage{booktabs} 
\usepackage{changepage,threeparttable}
\usepackage{makecell}
\usepackage{tablefootnote}

\newif\ifdraft
\definecolor{mygray}{gray}{0.7}
\definecolor{dkgreen}{RGB}{0,179,36}
\definecolor{dkorange}{RGB}{230,115,0}

\newcommand{\suwon}[1]{\ifdraft \textcolor{green}{[#1 --Suwon]} \fi}
\newcommand{\kl}[1]{\ifdraft \textcolor{blue}{[#1 --Karen]} \fi}
\newcommand{\ky}[1]{\ifdraft \textcolor{purple}{[#1 --KY]} \fi}

\ifdraft
\newcommand{\suwonadd}[1]{\textcolor{green}{{#1}}}
\newcommand{\suwondel}[1]{\textcolor{green}{{\st{#1}}}}
\newcommand{\kledit}[1]{\textcolor{blue}{{#1}}}
\newcommand{\klremove}[1]{\textcolor{blue}{{\st{#1}}}}
\newcommand{\kyedit}[1]{\textcolor{purple}{{#1}}}
\newcommand{\kyremove}[1]{\textcolor{purple}{{\st{#1}}}}
\else
\newcommand{\suwonadd}[1]{\textcolor{black}{{#1}}}
\newcommand{\suwondel}[1]{{}}
\newcommand{\kledit}[1]{\textcolor{black}{{#1}}}
\newcommand{\klremove}[1]{{}}
\newcommand{\kyedit}[1]{\textcolor{black}{{#1}}}
\newcommand{\kyremove}[1]{{}}
\fi

\title{Generative Context-aware Fine-tuning \\of Self-supervised Speech Models}

\name{Suwon Shon$^1$, Kwangyoun Kim$^1$, Prashant Sridhar$^1$, Yi-Te Hsu$^1$, Shinji Watanabe$^{2}$, Karen Livescu$^{3}$}
\address{$^1$ASAPP \ \ \ \ \   $^2$Carnegie Mellon University\ \ \ \ $^3$Toyota Technological Institute at Chicago }

\begin{document}
\ninept
\maketitle

\begin{abstract}

When performing tasks like automatic speech recognition or spoken language understanding for a given utterance, access to preceding text or audio provides contextual information can improve performance.
Considering the recent advances in generative large language models (LLM), we hypothesize that an LLM could generate useful context information using the preceding text.  
With appropriate prompts, LLM could generate a prediction of the next sentence or abstractive text like titles or topics. 
In this paper, we study the use of LLM-generated context information and propose an approach to distill the generated information during fine-tuning of self-supervised speech models, which we refer to as generative context-aware fine-tuning. 
This approach allows the fine-tuned model to make improved predictions without access to the true surrounding segments or to the LLM at inference time, while requiring only a very small additional context module. 
We evaluate the proposed approach using the SLUE and Libri-light benchmarks for several downstream tasks: automatic speech recognition, named entity recognition, and sentiment analysis. 
The results show that generative context-aware fine-tuning outperforms a context injection fine-tuning approach that accesses the ground-truth previous text, and is competitive with a generative context injection fine-tuning approach that requires the LLM at inference time.
% }

\end{abstract}

\section{Introduction}
\label{sec:intro}
% \kl{the first sentence could be removed -- I don't think it adds much} As the foundational pre-trained model becomes larger and more powerful, fine-tuning becomes crucial to direct the model towards a specific domain, task, and language depending on the downstream application.
Fine-tuning of speech models is typically performed at the utterance level, which means that context information from previous utterances is not taken into account. To address this issue, several approaches have been explored using preceding utterances or dialog history~\cite{hori2021advanced,tomashenko2020dialogue,wei2022conversational,wei2022leveraging,kim2019acoustic,kim2019cross}. 
% For short-term contextual information, many studies utilize adjacent audio frames or chunks~\cite{oord2018representation,tsunoo2019transformer,chung2019unsupervised,baevski2020wav2vec,kim2021multi,an2022cuside_IS,huybrechts23_interspeech}. \kl{these papers seem a bit off-topic to me, since we are mainly talking about context beyond the utterance level} 
% For longer-term contextual information, preceding utterances or dialog history are used

Recent studies show various ways to utilize previous audio and text history. 
\suwonadd{Cui et al.~\cite{cui23_interspeech}} \kledit{extract compact cross-utterance contextual features from the concatenated encoder output of preceding audio segments.}
%\kl{I can't parse the prev sentence}
\suwon{updated}\kl{edited a bit--it seemed too long}
Chang et al.~\cite{chang2023context} proposed an approach to inject a context embedding extracted from previous utterances using a pre-trained BERT model. 
This work is similar to the previous work proposed by Shon et al.~\cite{shon2023context}. The main difference is Shon et al. used the previous audio to extract context embedding so that the text-based language model is not needed.
While all these works are effective, there are several drawbacks. First, they need access to the previous utterances. Second, some of the works need an additional text-based language model whose size can be prohibitive during fine-tuning and inference.
% \kl{some of the prior work requires this access to previous utterances or to the language model during fine-tuning and some during inference.  It would be good to make that difference clearer.}

\begin{figure}[t]
    \centering
    \begin{subfigure}[b]{0.14\textwidth}
      \includegraphics[width=\textwidth]{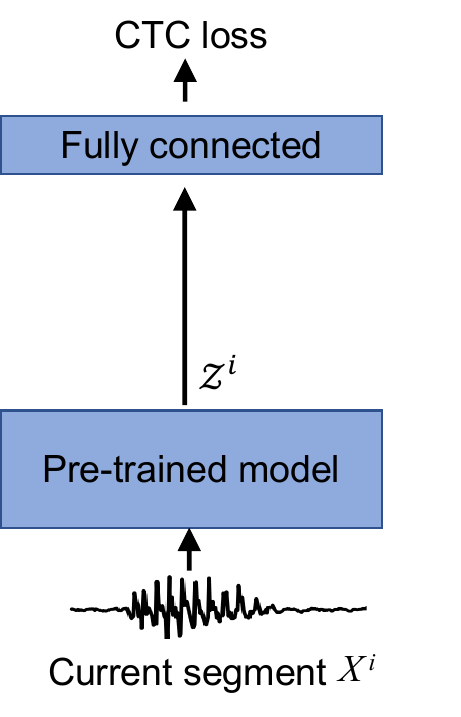}
         % \vspace{-0.5cm}
         \caption{}
    \end{subfigure}
    \hspace{0.3cm}
    \begin{subfigure}[b]{0.28\textwidth}
      \includegraphics[width=\textwidth]{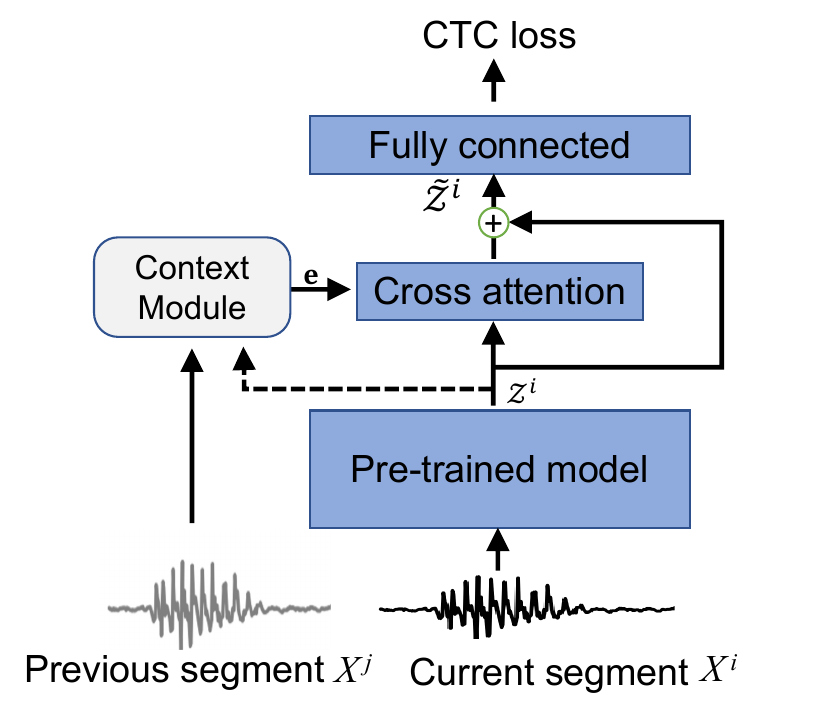}
         \caption{}
    \end{subfigure}
    \vspace{-0.2cm}
    \caption{Network structure overviews: (a) \kledit{B}aseline (b) \suwonadd{Proposed \kledit{approach} injecting \kledit{a} context embedding to \kledit{the} encoder output with cross attention.} The dotted line is required depending on the context module. 
    %\kl{is it correct to say this is the proposed approach?  this is more of a generic context-based model}\suwon{updated. I think it's part of proposal}
    }
    \label{fig:concept}
    \vspace{-0.3cm}
\end{figure}

In this paper, we propose {\it generative context-aware fine-tuning}, which utilizes \kledit{text generated by an LLM} to extract \kledit{a} context embedding. 
%\kl{need a few more words here to clarify what is "the generated text"}\suwon{updated}
Unlike previous studies that focus on techniques for context embedding extraction and integration into the model, we focus on how to generate useful context information using the preceding text.
With the latest progress in large language models (LLMs), we hypothesize that these models can generate useful context information by analyzing the preceding text. For example, by providing certain prompts, an LLM can anticipate the next reply in a conversation, considering the previous dialog history. Although the generated response may not be an exact match to the actual one, it can still help if it contains relevant information.
Our study explores the effectiveness of different prompts for candidates to provide context information using preceding utterances, using various sizes of LLMs. To extract the context embedding from the generated text, we used a pre-trained BERT model. Furthermore, we develop a generative context-aware fine-tuning module that distills the context embedding into a more compact additional network. This approach enables the system to infer contextual information without the need for previous utterances or any other language model prerequisites at inference time. This paper therefore makes two main contributions: (1) The study of multiple ways of generating contextual information, and (2) the distillation of the context module for more efficient inference.

\section{Generative Context-aware Fine-tuning}
\label{sec:generative}

\begin{figure*}[t]
    \centering
    \begin{subfigure}[b]{0.16\textwidth}
      \includegraphics[width=\textwidth]{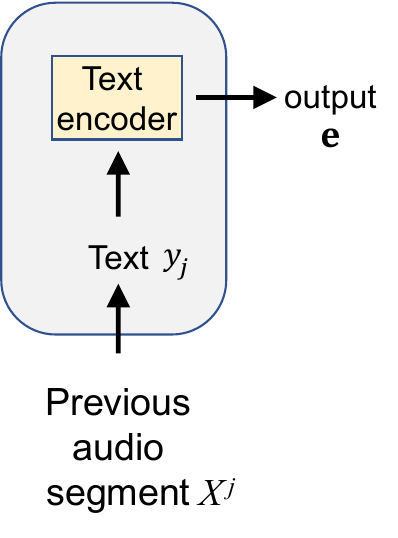}
         % \vspace{-0.5cm}
         \caption{Context injection}
         \label{subfig:concept_method}
    \end{subfigure}
    \begin{subfigure}[b]{0.33\textwidth}
      \includegraphics[width=\textwidth]{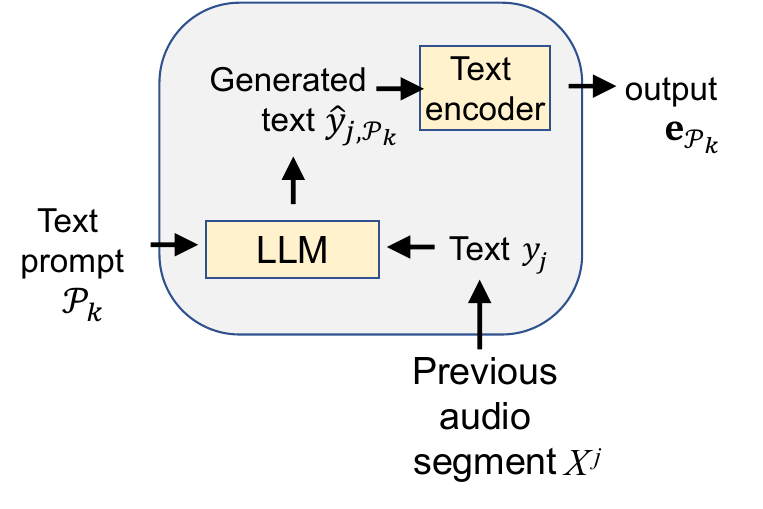}
         % \vspace{-0.5cm}
         \caption{Generative context injection}
    \end{subfigure}
    \begin{subfigure}[b]{0.48\textwidth}
      \includegraphics[width=\textwidth]{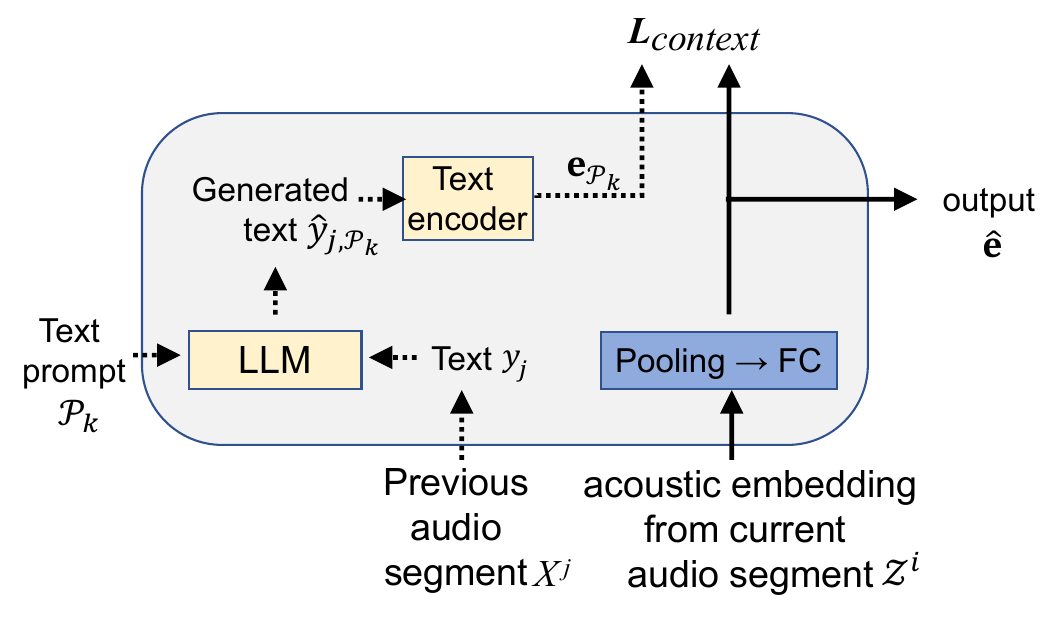}
         % \vspace{-0.5cm}
         \caption{Generative context-aware}
    \end{subfigure}

    \vspace{-0.1cm}
    \caption{Context module network structures. The dotted lines in (c) are not needed at inference time}
    \label{fig:context_modules}
    \vspace{-0.2cm}    
\end{figure*}

% \kl{I think this par can be deleted -- it is repeating material from the intro} Contextual information is crucial for accurate recognition when processing spoken language. This information can come in the form of a word or phrase or even higher-level knowledge (such as a topic of conversation) that has a semantic relationship in the speech that comes before or after the current segment.  While the self-attention layer of a transformer network is capable of capturing a wide range of context, the range is still limited by the current speech segment being processed. A previous study \ky{add reference} suggested that analyzing neighboring speech segments can capture a longer range of global context. 

Depending on the model architecture and application environment, the approach to incorporating context information can vary. However, the general concept involves injecting context embedding to the current audio embeddings. In this section, we demonstrate how we integrate a context module for self-supervised speech models and subsequently fine-tune/train models.

\subsection{Context module for speech models}

Fig.~\ref{fig:concept} (a) shows the high-level model architecture of our baseline context-free model. The pre-trained speech model generates speech representations $\mathcal{Z}^i = \{\mathbf{z}^i_1,\mathbf{z}^i_2,...,\mathbf{z}^i_{T_i}\}$ from input audio $\mathcal{X}^i$ where $i$ is the current segment index of the speech stream and $T_i$ is the frame length of $i$-th segment. The fully connected layer transforms the speech representations $\mathcal{Z}^i$ into embeddings of dimensionality equal to the target label vocabulary size, e.g.~the number of tokens, letters, or words. In the proposed approach with a context module, we use a cross-attention module $\mathtt{CA}$ to merge the context module output $\mathbf{e}$ and $\mathcal{Z}^i$, as shown \kledit{in} Fig.~\ref{fig:concept} (b). The new context-encoded speech representation is
\begin{equation} 
\Tilde{\mathcal{Z}}^i = \mathtt{CA}(\mathcal{Z}^i, \mathbf{e}) + \mathcal{Z}^i
\end{equation}

\kl{this section doesn't actually fully describe context-aware fine-tuning as the title suggests}\suwon{changed subsection title}

\subsection{Generative \kledit{context-aware} fine-tuning}

While previous studies mentioned in section~\ref{sec:intro} verified context information is effective on speech models, an important limitation is that the approaches rely on the previous audio or text to extract information. In this study, we hypothesize that a large-scale trained generative model could generate more useful information by consuming the previous speech/dialog history. 
% below can be shortened or removed. seems redundant.
We propose a generative context injection module to test if the generated text can enhance the speech model. Furthermore, we propose a generative context-aware module to enable the system to operate without the need for previous utterances or an LLM at inference time. 

\subsubsection{Context generation by prompting LLM}
To generate text using an LLM, a prompt is necessary and should be designed properly to obtain the desired output. We designed four prompts, shown in Table~\ref{tab:prompts}, that aim to extract contextual information from the given text of the previous utterance. These prompts correspond to the tasks of predicting the next sentence, a relevant question, the topic, and a title. Although the generated text seems reasonable based on the input text, there may be slight differences or misunderstandings when we compare it to the oracle following text.

\subsubsection{Generative context-aware module}

Motivated by a previous study~\cite{chang2023context}, we adapt their approach for fine-tuning self-supervised speech model scenarios as shown in Fig.~\ref{fig:context_modules} (a). At training time, we can use either the ground truth or automatically transcribed text $\mathbf{y}_{j}$ of the previous audio $\mathcal{X}^j$ where $j<i$, while at inference time we can only use automatically transcribed text. In this study, we fix $j$ as $i-1$.
% \ky{j is the index of previous segments? is j always i-1 or other case?}
The output $\mathbf{e}$ feeds into the cross-attention module in Fig.~\ref{fig:concept} (b). Since we are using a cross-attention module, the output $\mathbf{e}$ can be either a fixed-length embedding obtained by a pooling layer (e.g.classification ([CLS]) token) 
% \ky{what is the [CLS]?} 
or a variable-length sequence embedding (e.g. hidden states of the last layer).

\begin{table}[!t]\centering
\caption{An example ground-truth text of an audio segment in the slue-voxceleb dev set, and the result of prompting with our four prompts. A 13B parameter LLM\tablefootnote{stabilityai/StableBeluga-13B~\cite{StableBelugaModels}} was used in this example. The real title of this audio clip is ``my journey from marine to actor".}\label{tab:prompts}
\scriptsize
\vspace{-0.15cm}
\begin{tabular}{p{0.6in}p{2.36in}}\toprule
Input text (oracle) &living in the back room of my parents' house paying rent in the small town i was raised in northern indiana called mishawaka i could spell \\ 
following text (oracle) &and stories about of actors moving to la with like seven dollars and finding work and successful careers i got as far as amarillo texas before my car broke down i spent a \\\midrule
Prompt 1 ($\mathcal{P}_{1}$) &Provide a next sentence for the given text: \\
Generated text &\textit{it was a humble beginning, but it allowed me to save money and eventually pursue my dreams.} \\\midrule
Prompt 2 ($\mathcal{P}_{2}$)&This is part of the answer. Can you predict what was the question? text : \\
Generated text &\textit{what is your living situation and where are you from?} \\\midrule
Prompt 3 ($\mathcal{P}_{3}$) &Predict topic of the given text: \\
Generated text &\textit{the topic of the given text is living in the back room of my parents' house in mishawaka, indiana. }\\\midrule
Prompt 4 ($\mathcal{P}_{4}$)&Predict title of the given text: \\
Generated text &\textit{returning home navigating adulthood in mishawaka, indiana} \\
\bottomrule
\end{tabular}
\vspace{-0.4cm}
\end{table}

Fig.~\ref{fig:context_modules} (b) shows the proposed approach with generative context injection. The text encoder outputs $\mathbf{e}_{\mathcal{P}_{k}}$ by using the generated text $\hat{\mathbf{y}}_{j,\mathcal{P}_{k}}$ with the given previous text $\mathbf{y}_{j}$ and prompt $\mathcal{P}_{k}$. It is expected that the generated text $\hat{\mathbf{y}}_{j,\mathcal{P}_{k}}$ contains more useful context information than the text $\mathbf{y}_{j}$. However, there is a critical drawback:  This approach requires both an LLM and a text encoder at inference time. This introduces a significant increase in latency and computational cost.

To address this drawback, we propose a knowledge distillation method illustrated in Fig.~\ref{fig:context_modules} (c). In this approach, we create a similar context embedding $\hat{\mathbf{e}}$ by pooling and applying a fully connected layer to the acoustic embedding $\mathcal{Z}^i$ obtained from the current audio segment $\mathcal{X}^i$. The output $\mathbf{e}_{\mathcal{P}_{k}}$ from the text encoder is used solely in fine-tuning for computing an L2 loss between the new context embedding and the generative context embedding, as follows: 
\begin{equation}
    L_{context}=\|\mathbf{e}_{\mathcal{P}_{k}}-\hat{\mathbf{e}} \|_2.
\end{equation}
The context loss $L_{context}$ can be added to the loss for any downstream task with a weight $\alpha$.
This approach enables inference without requiring the LLM and text encoder, thereby significantly reducing model parameters compared to the generative context injection approach.

% - Describe 3 different approach.

% Previous studies showed surrounding text and audio could give context information while decoding current speech input and consequently lead to ASR performance improvement. Given this observation, we hypothesize that a large language model with proper prompt could generate better context information by leveraging knowledge learned from large-scale pre-training.

% Figure~\ref{fig:concept}. (a) shows text injection system. We can use ground truth text or transcribed text of previous audio segment during training.

% Figure~\ref{fig:concept}. (b) shows generated text injection system. We can generate text that may contain context information using LLM.

% Figure~\ref{fig:concept}. (c) shows generated text aware fine-tuning system. We can generate text that may contain context information using LLM. The context embedding vector can be learned during training to produce similar embedding to the text embedding from generated text.

\section{Experiments}
\label{sec:experiments}
\subsubsection{Tasks and datasets}

We conduct experiments on ASR and SLU downstream tasks using the \ky{Libri-light or Librispeech? I think other sections are using Librispeech instead. Train on Libri-light subset and Test on Librispeech?}Libri-light~\cite{kahn2020libri} and SLUE benchmarks~\cite{shon2022slue}. 
For ASR, we use SLUE-VoxCeleb and SLUE-VoxPopuli for the SLUE benchmark and use Librispeech~\cite{panayotov2015librispeech} for the Libri-light benchmark.
% For ASR, we use SLUE-VoxCeleb~\cite{shon2022slue,nagrani2017voxceleb}, SLUE-VoxPopuli~\cite{shon2022slue,wang2021voxpopuli}, and Libri-light~\cite{kahn2020libri}.
% We use SLUE-VoxCeleb and SLUE-VoxPopuli datasets (12.8 and 14.5 hours of audio, respectively) for fine-tuning.
For Libri-light, we fine-tune the models on different amounts of labeled audio (10m, 1h, 10h, and 100h).
% \kl{the prev few sentences are pretty repetitive, and a bit unclear about LibriSpeech vs. Libri-light}
For the SLU tasks, we use the SLUE-VoxPopuli dataset for named entity recognition (NER) and the SLUE-VoxCeleb dataset for sentiment analysis (SA).
NER involves detecting the named entities and their tags (types) in a given sentence. We evaluate an unordered list of named entity phrases and tag pairs predicted for each sentence using the F1 score.
SA classifies speech segments into negative, neutral, or positive sentiment, and is evaluated using a macro-averaged (unweighted) F1 score. We follow all pre-defined dataset splits and evaluation rules in each benchmark.

\subsection{Training setup}
The models are implemented using fairseq~\cite{ott2019fairseq} and SLUE-Toolkit~\cite{shon2022slue}. We use all of the hyper-parameters defined in SLUE-Toolkit for SLUE dataset fine-tuning and in fairseq for Librispeech fine-tuning, except for additional hyper-parameters added in the proposed approach. We use the pre-trained wav2vec2.0 base model as the pre-trained acoustic model for all experiments \suwonadd{since it is a robust acoustic model and also allows us to compare the performance to previous studies.}
\kl{it might be worth saying why wav2vec2.0, since it is typically outperformed by other models}\suwon{Added}
For the ASR and NER tasks, the target labels are characters plus word boundary tokens. The NER task has an additional 18 tokens to tag the start and end of each named entity. For the SA task, we replace CTC loss with cross-entropy loss, and the fully connected layer output dimension is 3 to produce sentiment class output. All fine-tuning is done 3 times with different random seeds, and we report the average performance. \kl{can you comment on the range of deviation from the average? (even though 3 is not a large number, it's good to know)}

For the text encoder, we use the BERT-base model. We consider both the [CLS] embedding and the last hidden layer output for comparison. For the text of the previous audio segment, we used the ground truth text.
% \kl{"considering the utterance ID ..." doesn't seem necessary}
For the LLM, we use multiple models varying in size from 7B to 70B parameters. All models are instruct\kledit{ion}-tuned \suwonadd{LLaMA~\cite{touvron2023llama,touvron2023llama2}}
\kl{say which LLM -- some LLaMA, right?} 
and we follow each model's prompt template. The generated text is limited to a maximum of 256 tokens with a greedy search decoding strategy. For cross-attention, we use a single head with 32 dimensions.

\subsection{Hyper-parameter optimization and ablation study}
\label{sec:hyper_exp}
Since there are multiple sub-modules in the proposed approach, we explore hyper-parameters to investigate the optimal settings. All hyper-parameter exploration was done using the 1h setting of the Libri-light benchmark.

\noindent\textbf{Baselines} Table~\ref{tab:baseline} shows baselines from previous studies, as well as the context injection approach (Fig.~\ref{fig:context_modules} (a)) with both [CLS] embedding and sequence embedding from the last hidden state. Compared to the audio-based context-aware system, the text-based context injection approach shows significant performance gain  (B vs. C1). [CLS] embedding performs slightly better than sequence embedding and is therefore used in all subsequent experiments. We also tested what happens if we feed a text embedding of the current decoded audio (C1 vs. C3 \suwonadd{and C2 vs. C4}). However, we found no significant performance difference, indicating that the contextualized word embedding (such as BERT embedding) does not directly aid ASR, even if \suwonadd{the oracle text of \kledit{the} input audio is fed to the text encoder to produce \kledit{the} context embedding.}\kl{I'm a bit confused by the last sentence.  Is the input ground-truth or decoded text?} \suwon{It's ground truth in this experiment. Rephrased.} \ky{What about clearly mentioning ground-truth text is used in this experiments? descriptions in Table 2 or early this part.}

\begin{table}[!t]\centering
\caption{LibriSpeech dev-other WER (\%) for baseline models with different embeddings types.}\label{tab:baseline}
\vspace{-0.15cm}
\begin{tabular}{l|lrrr}\toprule
&Model &Context type &WER \\\midrule
A &CTC (baseline)~\cite{baevski2020wav2vec} &- &29.6 \\
B &Context-aware~\cite{shon2023context}&Previous audio &28.8 \\
C1 &Context injection ([CLS]) &Previous text &25.2 \\
C2 &Context injection (sequence) &Previous text &25.5 \\
C3 &Context injection ([CLS]) & Current text &25.1 \\
C4 &Context injection (sequence) & Current text &25.2 \\
\bottomrule
\end{tabular}
\vspace{0.1cm}
\end{table}
\begin{table}[!t]\centering

\caption{ROUGE-1 F-score between generated text and current utterance text, and average number of words for several prompts. 
% \kl{"ROUGE-1" and "F-score" are formatted differently in different places in the paper.  I think these are standard but in any case it should be consistent.  Number columns should be aligned at the decimal point.  Other table captions could use a bit of grammar and formatting checking as well.}
}\label{tab:prompt_analysis}
% \begin{tabular}{p{1.25in}p{0.45in}p{0.4in}}\toprule
\vspace{-0.15cm}
\centering
\begin{tabular}{lrr}\toprule
Current GT text vs. &ROUGE-1 (F-score) &Average \# words \\\midrule
Previous GT text &0.14 &17.7 \\
Generated with $\mathcal{P}_{1}$ &0.12 &17.6 \\
Generated with $\mathcal{P}_{2}$ &0.09 &11.4 \\
Generated with $\mathcal{P}_{3}$ &0.08 &13.1 \\
Generated with $\mathcal{P}_{4}$ &0.06 &7.2 \\
\bottomrule
\end{tabular}
\vspace{0.1cm}
\end{table}

% \begin{table}[!htp]\centering
% \caption{Distance measure}\label{tab: }
% \begin{tabular}{l|rrrr}\toprule
% Current GT text Vs. &WER &Bertscore &rouge1 \\\midrule
% previous GT text &155.3\% &-0.02 &0.14 \\
% Generated with prompt 1 &156.2\% &0.01 &0.12 \\
% Generated with prompt 2 &125.1\% &-0.02 &0.09 \\
% Generated with prompt 3 &146.4\% &-0.07 &0.08 \\
% Generated with prompt 4 &104.1\% &-0.04 &0.06 \\
% \bottomrule
% \end{tabular}
% \end{table}

\begin{table}[!t]\centering
\caption{Librispeech dev-other WER on Generative context injection system (Fig. ~\ref{fig:context_modules}. (b)) by prompt type. All system used 13B instruct-tuned LLM}\label{tab:prompt_exp}
\vspace{-0.15cm}
\begin{tabular}{l|lrrr}\toprule
&Model &Prompt type &WER \\\midrule
C1 &Context injection ([CLS]) & - &25.2 \\
D1 &Generative context injection &$\mathcal{P}_{1}$ &25.1 \\
D2 &Generative context injection &$\mathcal{P}_{2}$ &24.9 \\
D3 &Generative context injection &$\mathcal{P}_{3}$ &25.1 \\
\textbf{D4} &Generative context injection &$\mathcal{P}_{4}$ &\textbf{24.9} \\
\bottomrule
\end{tabular}
\vspace{0.1cm}
\end{table}

\noindent
\textbf{Generative context injection:  Choice of prompt} The quality of the generated text is heavily dependent on the prompt. Prior to evaluating ASR task performance, we assess the ROUGE-1 F-score between the generated text 
% \ky{from what? ground-truth previous text?} 
and the current ground truth text in Table~\ref{tab:prompt_analysis}. As the Prompt $\mathcal{P}_{1}$ is for the next sentence prediction, it shows the highest ROUGE-1 F-score than the others. 
% \kl{grammar needs work in prev sentence} 
Additionally, we note that the text generated with $\mathcal{P}_{4}$ has the fewest words on average, which suggests that it is the most abstractive. 
% \kl{checked for tense up to this point.  In most cases you want to use present tense}
In the ASR task, we observed only a slight performance gap between the different prompts as shown in Table~\ref{tab:prompt_exp}.
We chose prompt $\mathcal{P}_{4}$ in other all experiments as it performed slightly better and generated shorter text compared to the other prompts. 
\suwonadd{Interestingly, the ROUGE-1 F-score of $\mathcal{P}_{1}$ is not greater than \kledit{that of} the previous GT text\klremove{,}\kledit{;} however, it is showing better ASR performance (C1 vs. D1).}
\kl{I think it would be good to add the context injection baseline for comparison to Table 4, and point out at this point that generative context already slightly improves over it.  
The most interesting thing to me from the Rouge-1 table is that P1 is not better than previous GT text, but is better in terms of downstream performance.  Rouge-1 is arguably not so meaningful for the other prompts, since they are not trying to generate anything like the next utt.}

\noindent
\textbf{LLM size}. In this experiment, we tested different LLM sizes in Table~\ref{tab:llm_size}. We hypothesized that a large LLM size could provide better context information. We observed that the context generated by the 7B LLM is worse than the ground truth text (D5 vs. C1). Increasing the LLM size from 7B to 13B resulted in a WER gain (D5 vs. D4). Consequently, we used the 13B LLM for all subsequent experiments, as we found no improvement beyond this size. \kl{this par could be shortened quite a bit if needed.  Also I would find it helpful to include C1 in Table 5.}
\suwon{C1 is added in the table4 and I think table 5 is in the same page, so should be good.}

\noindent
\textbf{Generative context-aware fine-tuning} 
% \kl{I suggest adding "Generative context-aware fine-tuning" to the par title, since that's the proposed approach}
Table~\ref{tab:distillation} shows that our proposed distillation with a simple pooling and fully connected student layer can effectively learn the \kyedit{context} embedding, with no degradation in WER, while significantly reducing the number of needed parameters at inference time. In these experiments we used the same hyper-parameters for the student part of the model (pooling and fully connected layer), the dimension and learning loss weight, as in prior work on context-aware fine-tuning~\cite{shon2023context}.

Finally, using all the hyper-parameters chosen in this section, we evaluate the systems on the Libri-light benchmark~\cite{kahn2020libri} and SLUE benchmark dataset\cite{shon2022slue}, with results shown in Table~\ref{tab:librilight_exp} and \ref{tab:slue_exp} respectively.

\subsection{Discussion}

\begin{table}[t!]\centering
\caption{Librispeech dev-other WER on Generative context injection system by LLM parameter size. All systems used the prompt 3 in Table~\ref{tab:prompts}. All LLMs are variant of llama-2~\cite{touvron2023llama2} except 30B model use llama~\cite{touvron2023llama}. All models are available at Huggingface platform}
\label{tab:llm_size}
\vspace{-0.15cm}
\begin{tabular}{l|lrrr}\toprule
&Model &LLM size &WER \\\midrule
D5 &Generative context injection &7B\tablefootnote{stabilityai/StableBeluga-7B~\cite{StableBelugaModels}} &26.1 \\
\textbf{D4} &Generative context injection &\textbf{13B}~\tablefootnote{stabilityai/StableBeluga-13B~\cite{StableBelugaModels}} &\textbf{24.9} \\
D6 &Generative context injection &30B~\tablefootnote{upstage/llama-30b-instruct-2048} &25.1 \\
D7 &Generative context injection &70B~\tablefootnote{stabilityai/StableBeluga2~\cite{StableBelugaModels}} &25.0 \\
\bottomrule
\end{tabular}
\end{table}

\begin{table}[!t]\centering
\caption{Librispeech dev-other WER with and without distillation layer performance comparison}\label{tab:distillation}
\vspace{-0.15cm}
\begin{tabular}{l|lrr}\toprule
&Model &WER \\\midrule
D4 &Generative context injection &24.9 \\
E &Generative context-aware &24.9 \\
\bottomrule
\end{tabular}
\end{table}

\begin{table}[!t]\centering
\caption{LibriSpeech dev/test set evaluation result when fine-tuning on Libri-light low-resource labeled data 
\kl{It would be great to have a bit longer and more memorable name than A, B, etc. if it's possible to make space.  I guess "9" means "9.0" etc?  it should be written that way.  is the bold face meant to indicate the best number in each setting?  (it is not always, and sometimes there are 2 identical best numbers).}\suwon{updated}
}\label{tab:librilight_exp}
\vspace{-0.15cm}
\begin{tabular}{l|rrrrr}\toprule
&dev-clean &dev-other &test-clean &test-other \\\midrule
& \multicolumn{2}{l}{\textbf{10min labeled}} & & \\
A &46.1 &51.5 &46.9 &50.9 \\
B &41.7 &48.7 &42.6 &49.0 \\
C1 &38.4 &46.1 &39.4 &45.9 \\
D4 &37.7 &45.1 &38.6 &45.1 \\
E &\textbf{36.3} &\textbf{43.5} &\textbf{37.2} &\textbf{43.5} \\ \midrule
& \multicolumn{2}{l}{\textbf{1h labeled}} & & \\
A &24.1 &29.6 &24.5 &29.7 \\
B &21.8 &28.8 &22.0 &29.4 \\
C1 &\textbf{18.0} &25.2 &18.5 &25.6 \\
D4 &\textbf{18.0} &\textbf{24.9} &\textbf{18.3} &\textbf{25.2} \\
E &\textbf{18.0} &\textbf{24.9} &\textbf{18.3} &25.3 \\ \midrule
& \multicolumn{2}{l}{\textbf{10h labeled}} & & \\
A &10.9 &17.4 &11.1 &17.6 \\
B &\textbf{9.0} &\textbf{16.4} &\textbf{9.1} &16.7 \\
C1 &9.6 &16.6 &9.7 &16.7 \\
D4 &9.3 &16.5 &9.3 &16.4 \\
E &9.3 &\textbf{16.4} &9.4 &\textbf{16.3} \\ \midrule
& \multicolumn{2}{l}{\textbf{100h labeled}} & & \\
A &6.1 &13.5 &6.1 &13.3 \\
B &\textbf{5.1} &\textbf{13.3} &\textbf{5.3} &13.0 \\
C1 &5.6 &13.5 &5.5 &12.9 \\
D4 &5.5 &13.5 &5.5 &12.8 \\
E &5.5 &\textbf{13.3} &5.4 &\textbf{12.7} \\
\bottomrule
\end{tabular}
\end{table}

\begin{table}[!t]\centering
\caption{SLUE benchmark scores on the dev/test set}\label{tab:slue_exp}
\vspace{-0.15cm}
\begin{tabular}{l|cc|c|c}\toprule
&\multicolumn{2}{c|}{ASR($\downarrow$)} &NER($\uparrow$) &SA($\uparrow$) \\\midrule
&VC &VP &VP &VC \\ \midrule
A &17.5 / 20.9 &17.5 / 18.4 &55.0 / 49.6 &49.2 / 51.8 \\
B &16.7 / 20.0 &16.4 / 17.0 &60.1 / 55.0 &49.7 / 52.9 \\
C1 &16.5 / 19.6 &16.9 / 17.8 &58.8 / 55.2 &49.0 / 52.8 \\
D4 &\textbf{16.2} / 19.3 &\textbf{16.1} / \textbf{16.9} &\textbf{60.9} / \textbf{56.8} &\textbf{51.7} / 52.2 \\
E &\textbf{16.2 / 19.2} &16.2 / 17.0 &59.7 / 56.5 &50.4 / \textbf{53.1} \\
\bottomrule
\end{tabular}
\end{table}

Based on the evaluation on the Libri-Light benchmark, we have observed that the proposed generative context-aware fine-tuning approach (E) is more effective when the fine-tuning dataset is limited. In particular, there is a 15\% relative improvement on the test-other set in the 10-minute setting and a 5\% increase in the 100-hour setting \suwonadd{compared to the baseline (A)}. \kl{relative to what?} It is also worth noting that our approach shows more improvement on noisy (``other") than on clean speech.\suwondel{This could be because the additional context embedding is learned from ground-truth text, not transcribed text.} \suwonadd{This indicates the noisy environment benefits from context more than a clean environment.}
\kl{that reasoning gave me pause when I read it.  why are we contrasting using ground-truth vs. not in training?  that doesn't seem to explain the increased benefit for noisy speech since both noisy and clean use ground-truth text.  isn't it just that there is more need for help from context when the speech is noisy?  also, it would be good to give percent improvements here too.}\suwon{correct. I think I forget to come back this part to revise.}In the evaluation on the SLUE benchmark, \suwonadd{we observed a similar performance gain in the ASR task as the Libri-light benchmark.}\suwondel{the ASR task showed a similar performance boost to that of the Libri-light benchmark.}\kl{grammar/wording needs a bit of work in prev sent}  While there is only slight improvement on the SA task, there was a significant improvement of 15\% on the NER task compared to the baseline \suwonadd{(A)}.\kl{which is the baseline here?} 
% These results suggest that a broader contextual information is advantageous for accurate dictation and tagging of named entity phrases by the proposed model.  \kl{the last sentence can be removed -- it is just restating the previous sentences}

It should be noted that the proposed module comprises several sub-modules, and despite having examined the hyper-parameters that had a notable effect on the performance, not all of them have been studied in this work. Further research is required to thoroughly investigate other design choices, such as the possibility of utilizing multiple previous utterances instead of just one as in this paper. Moreover, in addition to BERT, there are several other models we can consider for text encoding~\cite{he2021debertav3,raffel2020exploring,lewis2019bart}.

\section{Conclusion}
\label{sec:conclusions}
% \kl{edited and shortened conclusion.  previous version is in latex comment. BTW I suggest removing the "s" from the section title, since it is not about providing multiple conclusions.}
We have proposed a new approach for incorporating contextual information outside the current utterance, which is based on generating contextual text using a language model rather than relying on the existing signal context itself.  To our knowledge, this is the first effort to utilize LLMs in extracting context embeddings.  In addition, our distillation approach produces a model that performs just as well without requiring the actual context or LLM at inference time, only a very compact additional layer.  While our approach involves only the previous text, a natural next step is to combine it with prior context-aware fine-tuning work that uses the previous audio~\cite{shon2023context}.

%We have proposed a new approach that combines the context module with the self-supervised speech model. This approach demonstrates how context modules can be used in speech model architectures. Our experiments show that the proposed generative context-aware fine-tuning approach can improve performance with a very compact additional layer. To our knowledge, this is the first effort to utilize LLM in extracting context embedding. While our approach only involves the previous text, we can incorporate our prior study~\cite{shon2023context} that concentrates on the previous audio. By doing so, we can eventually integrate both preceding audio and text concurrently, which will maximize the exclusive information from either side. We plan to explore this further in our next study. 

\vfill\pagebreak
\clearpage

\bibliographystyle{IEEEbib}
{\bibliography{refs}}

\begin{thebibliography}{10}

\bibitem{hori2021advanced}
Takaaki Hori, Niko Moritz, Chiori Hori, and Jonathan~Le Roux,
\newblock ``Advanced long-context end-to-end speech recognition using
  context-expanded transformers,''
\newblock {\em arXiv preprint arXiv:2104.09426}, 2021.

\bibitem{tomashenko2020dialogue}
Natalia Tomashenko, Christian Raymond, Antoine Caubri{\`e}re, Renato De~Mori,
  and Yannick Est{\`e}ve,
\newblock ``Dialogue history integration into end-to-end signal-to-concept
  spoken language understanding systems,''
\newblock in {\em IEEE International Conference on Acoustics, Speech and Signal
  Processing (ICASSP)}, 2020, pp. 8509--8513.

\bibitem{wei2022conversational}
Kun Wei, Yike Zhang, Sining Sun, Lei Xie, and Long Ma,
\newblock ``Conversational speech recognition by learning conversation-level
  characteristics,''
\newblock in {\em IEEE International Conference on Acoustics, Speech and Signal
  Processing (ICASSP)}, 2022, pp. 6752--6756.

\bibitem{wei2022leveraging}
Kun Wei, Yike Zhang, Sining Sun, Lei Xie, and Long Ma,
\newblock ``Leveraging acoustic contextual representation by audio-textual
  cross-modal learning for conversational asr,''
\newblock {\em arXiv preprint arXiv:2207.01039}, 2022.

\bibitem{kim2019acoustic}
Suyoun Kim and Florian Metze,
\newblock ``Acoustic-to-word models with conversational context information,''
\newblock {\em arXiv preprint arXiv:1905.08796}, 2019.

\bibitem{kim2019cross}
Suyoun Kim, Siddharth Dalmia, and Florian Metze,
\newblock ``Cross-attention end-to-end asr for two-party conversations,''
\newblock {\em arXiv preprint arXiv:1907.10726}, 2019.

\bibitem{cui23_interspeech}
Mingyu Cui, Jiawen Kang, Jiajun Deng, Xi~Yin, Yutao Xie, Xie Chen, and Xunying
  Liu,
\newblock ``{Towards Effective and Compact Contextual Representation for
  Conformer Transducer Speech Recognition Systems},''
\newblock in {\em Proc. INTERSPEECH 2023}, 2023, pp. 2223--2227.

\bibitem{chang2023context}
Shuo-Yiin Chang, Chao Zhang, Tara~N Sainath, Bo~Li, and Trevor Strohman,
\newblock ``Context-aware end-to-end asr using self-attentive embedding and
  tensor fusion,''
\newblock in {\em IEEE International Conference on Acoustics, Speech and Signal
  Processing (ICASSP)}, 2023.

\bibitem{shon2023context}
Suwon Shon, Felix Wu, Kwangyoun Kim, Prashant Sridhar, Karen Livescu, and
  Shinji Watanabe,
\newblock ``Context-aware fine-tuning of self-supervised speech models,''
\newblock in {\em IEEE International Conference on Acoustics, Speech and Signal
  Processing (ICASSP)}, 2023.

\bibitem{StableBelugaModels}
Dakota Mahan, Ryan Carlow, Louis Castricato, Nathan Cooper, and Christian
  Laforte,
\newblock ``Stable beluga models,'' .

\bibitem{kahn2020libri}
J.~Kahn, M.~Riviere, W.~Zheng, E.~Kharitonov, Q.~Xu, P.~E. Mazare, J.~Karadayi,
  V.~Liptchinsky, R.~Collobert, C.~Fuegen, T.~Likhomanenko, G.~Synnaeve,
  A.~Joulin, A.~Mohamed, and E.~Dupoux,
\newblock ``{Libri-Light: A Benchmark for ASR with Limited or No
  Supervision},''
\newblock in {\em IEEE International Conference on Acoustics, Speech and Signal
  Processing (ICASSP)}, 2020.

\bibitem{shon2022slue}
Suwon Shon, Ankita Pasad, Felix Wu, Pablo Brusco, Yoav Artzi, Karen Livescu,
  and Kyu~J Han,
\newblock ``Slue: New benchmark tasks for spoken language understanding
  evaluation on natural speech,''
\newblock in {\em IEEE International Conference on Acoustics, Speech and Signal
  Processing (ICASSP)}. IEEE, 2022, pp. 7927--7931.

\bibitem{panayotov2015librispeech}
Vassil Panayotov, Guoguo Chen, Daniel Povey, and Sanjeev Khudanpur,
\newblock ``{Librispeech: An ASR corpus based on public domain audio books},''
\newblock in {\em IEEE International Conference on Acoustics, Speech and Signal
  Processing (ICASSP)}, 2015.

\bibitem{ott2019fairseq}
Myle Ott, Sergey Edunov, Alexei Baevski, Angela Fan, Sam Gross, Nathan Ng,
  David Grangier, and Michael Auli,
\newblock ``fairseq: A fast, extensible toolkit for sequence modeling,''
\newblock in {\em NAACL demo}, 2019.

\bibitem{touvron2023llama}
Hugo Touvron, Thibaut Lavril, Gautier Izacard, Xavier Martinet, Marie-Anne
  Lachaux, Timoth{\'e}e Lacroix, Baptiste Rozi{\`e}re, Naman Goyal, Eric
  Hambro, Faisal Azhar, et~al.,
\newblock ``Llama: Open and efficient foundation language models,''
\newblock {\em arXiv preprint arXiv:2302.13971}, 2023.

\bibitem{touvron2023llama2}
Hugo Touvron, Louis Martin, Kevin Stone, Peter Albert, Amjad Almahairi, Yasmine
  Babaei, Nikolay Bashlykov, Soumya Batra, Prajjwal Bhargava, Shruti Bhosale,
  et~al.,
\newblock ``Llama 2: Open foundation and fine-tuned chat models,''
\newblock {\em arXiv preprint arXiv:2307.09288}, 2023.

\bibitem{baevski2020wav2vec}
Alexei Baevski, Yuhao Zhou, Abdelrahman Mohamed, and Michael Auli,
\newblock ``wav2vec 2.0: A framework for self-supervised learning of speech
  representations,''
\newblock {\em NeurIPS}, 2020.

\bibitem{he2021debertav3}
Pengcheng He, Jianfeng Gao, and Weizhu Chen,
\newblock ``Debertav3: Improving deberta using electra-style pre-training with
  gradient-disentangled embedding sharing,''
\newblock {\em arXiv preprint arXiv:2111.09543}, 2021.

\bibitem{raffel2020exploring}
Colin Raffel, Noam Shazeer, Adam Roberts, Katherine Lee, Sharan Narang, Michael
  Matena, Yanqi Zhou, Wei Li, and Peter~J Liu,
\newblock ``Exploring the limits of transfer learning with a unified
  text-to-text transformer,''
\newblock {\em The Journal of Machine Learning Research}, vol. 21, no. 1, pp.
  5485--5551, 2020.

\bibitem{lewis2019bart}
Mike Lewis, Yinhan Liu, Naman Goyal, Marjan Ghazvininejad, Abdelrahman Mohamed,
  Omer Levy, Ves Stoyanov, and Luke Zettlemoyer,
\newblock ``Bart: Denoising sequence-to-sequence pre-training for natural
  language generation, translation, and comprehension,''
\newblock {\em arXiv preprint arXiv:1910.13461}, 2019.

\end{thebibliography}
% {\footnotesize \bibliography{refs}}

\end{document}